\documentclass[10pt,twocolumn,letterpaper]{article}

\usepackage[pagenumbers]{wacv} 

\usepackage[pagebackref,breaklinks,colorlinks]{hyperref}      
\usepackage{url}            
\usepackage{booktabs}       
\usepackage{amsfonts}       
\usepackage{nicefrac}       
\usepackage{microtype}      
\usepackage{graphicx}
\usepackage[table]{xcolor}
\usepackage{amsmath}
\usepackage{bbm}
\usepackage{enumitem}
\usepackage{amssymb}
\usepackage[linesnumbered,ruled]{algorithm2e}
\usepackage{mathabx}
\usepackage{subcaption}
\usepackage{siunitx}
\usepackage{booktabs, makecell, multirow}

\usepackage{wrapfig}

\usepackage{pifont}
\newcommand{\cmark}{\ding{51}}
\newcommand{\xmark}{\ding{55}}
  
\usepackage{url}
\usepackage[capitalize]{cleveref}
\crefname{section}{Sec.}{Secs.}
\Crefname{section}{Section}{Sections}
\Crefname{table}{Table}{Tables}
\crefname{table}{Tab.}{Tabs.}

\def\wacvPaperID{1513} 
\def\confName{WACV}
\def\confYear{2025}

\begin{document}

\title{Socially-Informed Reconstruction for Pedestrian Trajectory Forecasting}

\author{Haleh Damirchi \hspace{8mm} Ali Etemad \hspace{8mm} Michael Greenspan\\
Dept. ECE \& Ingenuity Labs Research Institute\\
Queen’s University, Kingston, Canada\\
{\tt\small {haleh.damirchi,ali.etemad,michael.greenspan}@queensu.ca}
}

\maketitle


\begin{abstract}

Pedestrian trajectory prediction remains a challenge for autonomous systems, particularly due to the intricate dynamics of social interactions. Accurate forecasting requires a comprehensive understanding not only of each pedestrian's previous trajectory but also of their interaction with the surrounding environment, an important part of which are other pedestrians moving dynamically in the scene. To learn effective socially-informed representations, we propose 
a model that uses a reconstructor alongside a conditional variational autoencoder-based trajectory forecasting module. This module generates pseudo-trajectories, which we use as augmentations throughout the training process. To further guide the model towards social awareness, we propose a novel social loss that aids in forecasting of more stable trajectories. We validate our approach through extensive experiments, demonstrating strong performances in comparison to state-of-the-art methods on the ETH/UCY and SDD benchmarks.

\end{abstract}

\section{Introduction}

Pedestrian trajectory prediction estimates a target pedestrian's future location after examining the observed locations of the pedestrian in the scene \cite{sociallstm}. This capability is essential for safe route planning of autonomous vehicles~\cite{autonomous} as it requires accurate and reliable predictions to ensure the safety of pedestrians and drivers~\cite{onboard}. By accurately forecasting future paths of pedestrians in the scene, proactive measures can be taken to prevent potential accidents~\cite{survey}. Similarly, trajectory forecasting helps to understand and model pedestrian behaviour, allowing  vehicles to anticipate and react to pedestrians' movements in real time~\cite{contextaware, bitrap, pedformer}. Additionally, it can assist in the analysis of pedestrian movement patterns in crowded areas to optimize pedestrian infrastructure~\cite{anomaly}.

A key aspect of human trajectory forecasting that distinguishes it apart from other time-series problems \cite{Autoformer,shomeecg} is the inherent human element \cite{behavior}. Humans are social beings, and interactions between individuals significantly influence how we navigate through (especially crowded) spaces \cite{sociallstm}. For instance, when walking in a crowded area, people naturally adjust their paths to avoid potential collisions with others. This social dynamic is critical in predicting human movements accurately. Prior works \cite{agentformer,socialgan,socialimplicit,socialstgcnn} have shown that considering social interactions in trajectory forecasting models can lead to more precise predictions, as these interactions play a crucial role in determining individual movement patterns.
Another challenge in human trajectory forecasting, which is shared by many deep learning applications, is the high cost of collecting labeled data. A standard approach to mitigate this issue is to apply standard augmentations to the data prior to training \cite{textaug,imaug1,imaug2}. However, while standard augmentation techniques have been initially developed and are well-suited for static images, they do not necessarily translate well to trajectory data. As a result, developing methods to generate more effective augmented trajectories, particularly those that consider social interactions, can significantly enhance the learning process, leading to better performance in real-world applications.

\begin{figure*}[t!]
\centerline{\includegraphics[width=0.9\linewidth]
{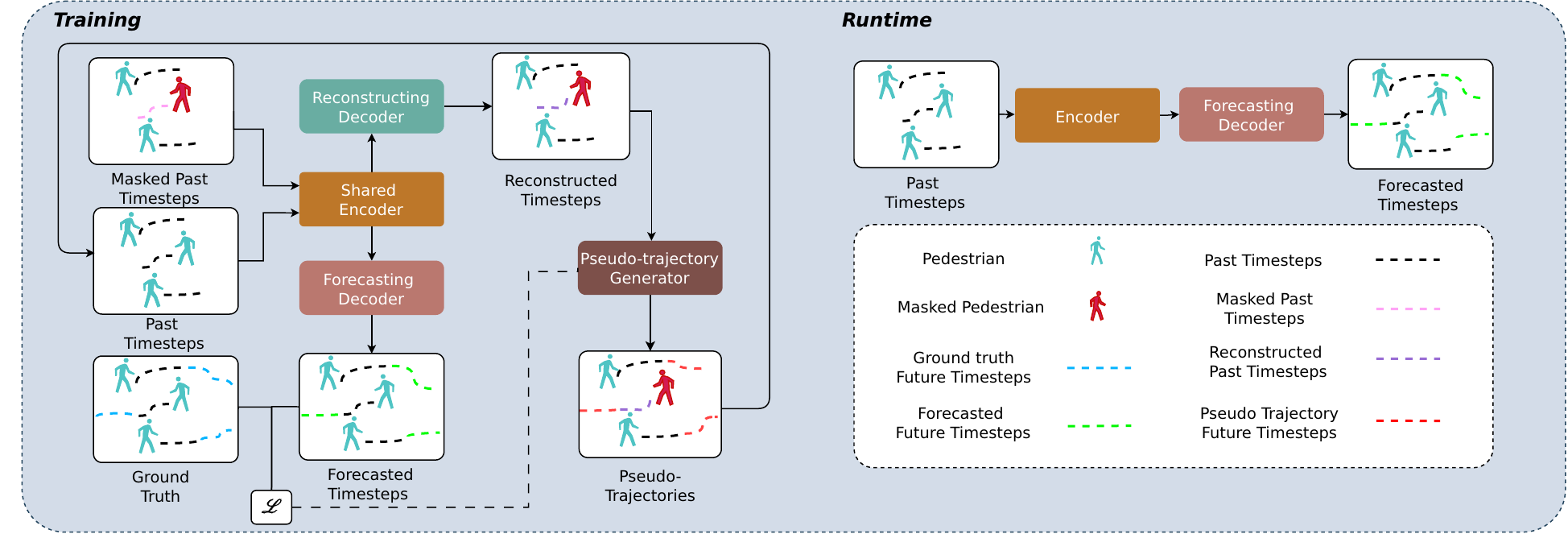}}
\caption{Overview of our proposed method. During training we continuously generate new socially-aware samples and add them to the training set. At runtime, we only use the encoder and forecaster to predict future timesteps.}
\label{fig:overview}
\end{figure*}

In this paper, to address the problems described above, we propose a novel trajectory forecasting model. Our method uses several distinct elements to improve representations and enhance the stability of the predictions. First, it uses a trajectory reconstruction module to operate alongside the forecaster. The inclusion of this module improves the learned representations and simultaneously allows the reconstructed pseudo-trajectories to be fed back to the training pipeline as augmentations. 
Our model only selects challenging pseudo-trajectories to be used as augmentations to ensure that the model is exposed to difficult scenarios that improve its robustness and generalization capabilities. Second, our model includes a novel loss, called social loss, to enforce socially accurate predictions in the generated future time-stamps. 
We illustrate an overview of our method in Figure~\ref{fig:overview}.
To evaluate the performance of our solution, we experiment on 5 popular pedestrian trajectory forecasting benchmarks \cite{eth, ucy}, namely ETH, Hotel, Univ, Zara1, Zara2, and Stanford Drone Dataset (SDD)~\cite{sdd}. Our experiments show our method to outperform the state-of-the-art based on standard forecasting metrics, while showing more stable performance across all predictions (as opposed to only the top prediction), especially as a result of our social loss. Various ablation and sensitivity studies demonstrate the impact of different components of our method.

In summary, we make the following contributions.
(\textbf{1}) We propose a novel solution for human trajectory forecasting. Our method uses a reconstructor alongside the forecasting module to learn stronger trajectory representations as well as to generate pseudo-trajectories, the more difficult of which is used as augmentations throughout the training process.
(\textbf{2}) We propose the use of a novel loss to enforce physically and socially viable trajectories, which aids in generating more stable predictions.
(\textbf{3}) We perform experiments on the widely utilized benchmarks and demonstrate that our method outperforms state-of-the-art trajectory prediction methods. We make our code publicly available at \url{https://github.com/thisishale/SocRec}.


\section{Related Work}  

\subsection{Trajectory Forecasting}
A variety of methods have focused on the social attribute of trajectories, in which multiple pedestrians are influenced by each other. Social GAN~\cite{socialgan} employed adversarial training to improve the accuracy of their model in complicated social scenarios. By synthesizing realistic trajectories, it refines predictions by incorporating both individual and social context. A pooling module was used to aggregate features from other pedestrians in the scene, though this module could cause the model to struggle in crowded environments.
Later, Trajectron++~\cite{trajectron++} introduced a social model for trajectory forecasting, where interactions were modelled by summing the feature vectors of neighboring pedestrians. While computationally efficient and a relatively simple operation, this method could oversimplify the interactions in a scene.
Social Spatio-Temporal Graph Convolutional Neural Network (Social-STGCNN)~\cite{socialstgcnn} departs from previous aggregation methods and uses a graph network with an adjacency matrix based on distances between pedestrians
to forecast future trajectories. Although this operation allows the model to account for neighbouring pedestrians, it may not adapt to changing social contexts such as group behaviors due to the non-learnable adjacency matrix.
To improve the social connection between trajectories, Agentformer~\cite{agentformer} proposed a new form of attention in their transformer architecture.
Whereas methods prior to Agentformer generally modelled the social and temporal information separately, the new form of attention jointly addressed both the temporal and the social aspects of trajectory prediction. Later, Social Implicit~\cite{socialimplicit} proposed to forecast each pedestrian's future trajectory by using sub-modules trained for specific speed ranges. This approach may not fully capture scenarios where pedestrians adjust their speed in response to others.

A different set of methods explore the importance of goals, i.e. the final timestep, for trajectory prediction~\cite{tnt,notthejourney,bitrap}. Unlike previous models that focus on a single, long term goal, SGNet~\cite{sgnet} introduced a 
step-wise goal estimator which dynamically estimated and utilizes goals at multiple timesteps. YNet~\cite{ynet} cast  goal estimation and future trajectory prediction into image form, utilizing convolutional neural networks in the form of a UNet architecture to predict both the goals and the future trajectory by sampling from the generated heatmap at the network output. However, both SGNet and YNet lack a structure for the social prediction of trajectories.

\subsection{Social, Psychological, and Behavioral Factors}
Incorporating environmental context, psychological factors and behaviour patterns of pedestrians is crucial for  predicting trajectories and motions of humans. Psychological and behavior attributes such as personal space and goal-oriented behaviour is explored and simulated in~\cite{simulation}. In this study a simulation is implemented where pedestrians adjust their trajectories based on attractive and repulsive forces in their environment, which reflects psychological tendencies. They conclude that pedestrians experience \textit{motivations to act}, which drives them to accelerate or decelerate as a reaction to the perceived information from their environment. In~\cite{walkalone}, an extended Kalman filter is used to create a dynamic model to consider the social interactions between pedestrians for motion tracking. They also use contextual information such as pedestrian heading to estimate the destination of pedestrains. Additionally, pedestrian walking style is explored in~\cite{etiquette}, where pedestrian trajectories are categorized into clusters based on their \textit{social sensitivity}. A low social sensitivity value indicates that a pedestrian's navigation is not dependant on other targets in the scene. Each cluster is evaluated in a separate model that is specialized for its corresponding category. Pedestrian intention has also been used in previous works for pedestrian trajectory prediction from a vehicle's view point~\cite{pie}. In this study, cropped images of pedestrians and their bounding box locations on dashcam images are fed to a recurrent neural network, and the future locations of bounding boxes on the image are estimated.

\section{Method}

\subsection{Problem Setup} \label{sec:problem}

Let $X_{t}\!\!=\!\!\{x^{1}_{t}, x^{2}_{t}, ..., x^{N}_{t}\}$ represent a set of 2D locations of $N$ pedestrians at time $t$, where $x_t^i\in \mathbb{R}^2$. 
We denote $S_{p}\!\!=\!\!\{X_{1}, X_{2}, ..., X_{t_{p}}\}$ and $S_{f}\!\!=\!\!\{X_{t_{p}+1}, X_{t_{p}+2}, ..., X_{t_{f}}\}$ as sequences of locations of $N$ pedestrians at the past and future timesteps, respectively, with $t_{p}$ and $t_{f}$ as the present and final timesteps. The goal of social trajectory forecasting is to accurately predict multiple plausible future trajectories given $S_{p}$, such that $x^i_t - x^j_t > \epsilon$, $\forall \ i \neq j$,
where $\epsilon$ is the minimum acceptable distance between two pedestrians at a given timestep.

\subsection{Proposed Approach}

\begin{figure*}[t!]
\centerline{\includegraphics[width=0.95\textwidth]{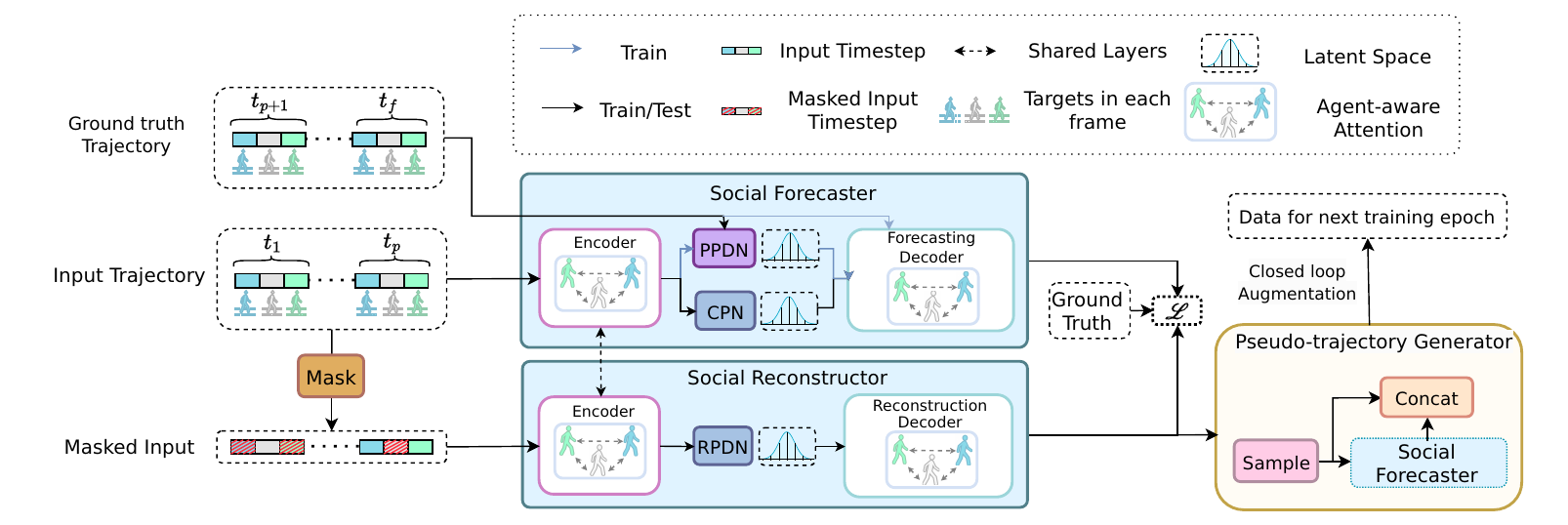}}
\caption{A detailed depiction of our method is presented. Our proposed model uses a social forecaster, a social reconstructor, and a pseudo-trajectory generator to augment the training data. The loss shown here consists of the forecaster loss, the reconstructor loss, and our novel social loss. 
}
\label{fig:method}
\end{figure*}

To address the above-mentioned problem, we propose a new solution which consists of three key components: a social forecaster, a social reconstructor, and a pseudo-trajectory generator. At a high level, the social forecaster predicts the future trajectory locations for each pedestrian in the scene given their locations in past timesteps. The social reconstructor reproduces the locations at past timesteps which have been partially masked for each pedestrian. Finally, the pseudo-trajectory generator creates new trajectory sequences by sampling from the reconstructed trajectories, which will then be used in future training iterations. The architecture for our proposed approach is illustrated in Figure~\ref{fig:method}. Following, we describe each element of our method in detail.

\subsubsection{Social Forecaster} 

The past trajectories $S_{p}$ are first fed to the social forecaster module, which is inspired by the 
Conditional Variational Autoencoder-based forecaster proposed in~\cite{agentformer} and consists of four components: Encoder, Conditional Prior Network (CPN), Predictor Posterior Distribution Network (PPDN), and Forecasting Decoder. The encoder module consists of a transformer encoder which uses 
agent-aware attention~\cite{agentformer}. 
The input to this attention unit is query $Q$, key $K$ and value $V$. The attention output is calculated as:
\begin{equation}
\small
\begin{aligned}
SocialAtt(Q,K,V)=softmax(\dfrac{A}{\sqrt{d_{m}}})V,
\label{eq:att1}
\end{aligned}
\end{equation}
\begin{equation}
\small
\begin{aligned}
A=M \odot (Q_{self}K_{self}^T)+(1-M) \odot (Q_{other}K_{other}),
\label{eq:att2}
\end{aligned}
\end{equation}
where $A$ is the attention weight matrix; $d_{m}$ is the model dimension; $Q_{self}\!=\!QW^Q_{self}$ and $Q_{other}\!\!=\!\!QW^Q_{other}$
are the query values projected by weights $W^Q_{self}$ and $W^Q_{other}$, and;
$K_{self}\!=\!KW^K_{self}$
and $K_{other}\!=\!KW^K_{other}$ 
are the key values projected by weights $W^K_{self}$ and $W^K_{other}$.
In Eq.~\ref{eq:att2}, for matrix $M$, the attention mask
between pedestrian $i$ and $j$ is defined as $M_{ij}\!=\!\mathbb{I}(i \text{ mod } N\!=\!j \text{ mod } N)$. When $i\!\!=\!\!j$, the attention weight matrix $A$ is masked to include only those weights that correspond to the relationship between the pedestrian itself at different timesteps. However, when $i\!\neq\!j$, the attention weight matrix represents the relationship between pedestrian $i$ and pedestrian $j$ at different timesteps.

To produce the latent code $z\!\in\!\mathbb{R}^{N \times d_{z}}$ for all $N$ target pedestrians, we pass the encoded features to the CPN module, which is a linear layer. This module generates the parameters $\mu_{i}^{p_{f}}$ and $\sigma_{i}^{p_{f}}$ for the prior Normal distribution $p_{{\phi}_{F}}(z|S_{p})=\mathcal{N}(\mu_{i}^{p_{f}},Diag(\sigma_{i}^{p_{f}})^2)$, where $\phi_F$ is the set of parameters for the CPN module. PPDN consists of an agent-aware cross-attention mechanism (explained earlier), followed by a linear layer similar to that of the CPN. This module utilizes both ground truth (future) and past timesteps to output the parameters of the posterior distribution $q(z|S_{p},S_{f})=\mathcal{N}(\mu_{i}^{q_f},Diag(\sigma_{i}^{q_f})^2)$. The latent code generated by PPDN and the ground truth future timesteps are then passed on to the transformer forecasting decoder at training time. This decoder consists of an agent-aware self-attention module followed by cross-attention. Masking is implemented in the agent-aware self-attention inside the decoder to prevent the previous timesteps from peeking into the future ground truth values.

To train the social forecaster, the negative evidence lower bound (ELBO) is employed as follows:
\begin{equation}
\small
\begin{aligned}
L_{F} = -\mathbb{E}_{{q_{\phi_F}}(z|S_{p},S_{f})}[log(p_{{\theta}_{F}}(S_{f}|z,S_{p}))]\\
+ KL(q_{{\phi}_{F}}(z|S_{p},S_{f})||p_{{\phi}_{F}}(z|S_{p})),
\label{eq:lp1}
\end{aligned}
\end{equation}
where $p_{{\theta}_{F}}(S_{f}|z,S_{p})$ is the conditional likelihood. The initial term in the formula denotes the prediction error, for which the Mean Squared Error (MSE) between the predicted future timesteps and the ground truth is utilized,
resulting in the following reformulation:
\begin{equation}
\small
L_{F} = \text{\textit{MSE}}(\tilde{S_{f}},S_{f})+KL(q_{{\phi}_{F}}(z|S_{p},S_{f})||p_{{\phi}_{F}}(z|S_{p})),
\end{equation}
where $\tilde{S_{f}}$ is the forecasted trajectory.
At training, the loss function minimizes the divergence between the posterior $q(z|S_{p},S_{f})$ and conditional prior distributions $p_{{\phi}_{F}}(z|S_{p})$, while ensuring the accuracy of the predictions. However, at test time, ground truth future sequences $S_{f}$ are not available. Thus, PPDN is only used during training to condition the predictions on the distribution of future timesteps.

To guide the social forecaster module towards avoiding overlap between predicted pedestrian locations, we introduce a novel loss term that penalizes the model for forecasting future pedestrian locations that are closer to each other than a threshold $\epsilon$. The proposed loss function is given by 
\begin{equation}
\small
\begin{aligned}
L_{SocF}\!=\!\dfrac{1}{m}\sum\limits_{t=t_p}^{t_f}\sum\limits_{i=1}^{N}\sum\limits_{j=i}^{N}
max(0,\epsilon\!-\!
dist(\tilde{S}_f(i,t)
,\tilde{S}_f(j,t))),
\end{aligned}
\label{eq:socf}
\end{equation}
where $dist(\tilde{S}_f(i,t)
,\tilde{S}_f(j,t)))$ is the squared euclidean distance between locations of pedestrians $i$ and $j$ at timestep $t$ and $m=N(N\!\!-\!\!1)/2$.

\subsubsection{Social Reconstructor} 
To further enforce the training process and provide stronger encoded representations of past trajectories $S_P$, our model includes a social reconstructor module in parallel with the social forecaster, where the encoder between the two modules are shared. To create the input for the social reconstructor, we mask the past trajectory $S_p$ by a ratio of $R$. We zero out $R\%$ of the total timesteps in the trajectory selected randomly, which we name $S^{\text{masked}}_{p}$. The social reconstructor module is designed to fill in the missing points in $S^{\text{masked}}_{p}$ utilizing a Variational Autoencoder (VAE). 
This module consists of an Encoder, a Reconstructor Posterior Distribution Network (RPDN), and a Reconstruction Decoder. The masked input $S^{\text{masked}}_{p}$ is first passed through the encoder module, which is shared with the social forecaster module. Subsequently, the RPDN, which is a linear layer, estimates the parameters $\mu_{i}^{q_r}$ and $\sigma_{i}^{q_r}$ for the posterior distribution $q(z|S_{p})=\mathcal{N}(\mu_{i}^{q_r},Diag(\sigma_{i}^{q_r})^2)$ of the encoded representations. The sampled latent code obtained from this module is then passed through the reconstruction decoder, which is a transformer decoder that uses the agent-aware attention described earlier.

The loss function to train the reconstructor module is:
\begin{equation}
\small
L_{R} = -\mathbb{E}_{{q_{\phi}}_{r}(z|S_{p})}logp_{{\theta}_{r}}(S_{p}|z)+KL(q_{{\phi}_{r}}(z|S_{p})||p_{{\phi}_{r}}(z)).
\end{equation}
Here, ${\phi}_{r}$ is the set of parameters for the RPDN and $p_{{\phi}_{r}}(z)$ is a Gaussian distribution with mean of 0 and variance of 1. Similar to Eq.~\ref{eq:lp1}, the first term is replaced with the MSE function as
\begin{equation}
\small
L_{R} = \textit{\text{MSE}}(\tilde{S_{p}},S_{p})+KL(q_{{\phi}_{r}}(z|S_{p})||p_{{\phi}_{r}}(z)),
\end{equation}
where $\textit{\text{MSE}}(\tilde{S_{p}},S_{p})$ is the Mean Squared Error between the reconstructed sequence $\tilde{S_{p}}$ and the observed sequences. 

Similar to the forecaster module, we propose to use a social loss for the reconstruction module to reconstruct the masked sequences while avoiding overlaps with other pedestrians in the scene as follows:
\begin{equation}
\small
\begin{aligned}
L_{SocR}\!=\!\dfrac{1}{m}\sum\limits_{t=1}^{t_p}\sum\limits_{i=1}^{N}\sum\limits_{j=i}^{N}
max(0,\epsilon\!-\!
dist(\tilde{S}_p(i,t)
,\tilde{S}_p(j,t))).
\end{aligned}
\label{eq:socr}
\end{equation}

\subsubsection{Total Loss}
The final loss function is formulated as:
\begin{equation}
\small
\begin{aligned}
\mathcal{L}_{Total} = w_{1}L_F+w_{2}L_R+w_3(L_{SocF}+L_{SocR}),
\end{aligned}
\end{equation}
where $w_{1}$, $w_{2}$ are corresponding weights for the CVAE, and VAE (reconstructor) modules, and $w_{3}$ is the weight for the social loss functions for both the forecaster and reconstructor. 

\subsubsection{Pseudo-trajectory Generation} 

We propose a pseudo-trajectory generator for augmenting the training set with samples that are challenging for the forecaster module. 
We follow the strategy proposed in \cite{rebalancingstrategy} and monitor the fluctuations in the loss value for each sample. Throughout training, each sample goes through phases of `descending' where the loss decreases for that sample between two consecutive epochs, and `ascending' where it increases. Accordingly, difficult samples tend to experience more `ascending' states than `descending' throughout the training. 
As the approach introduced in~\cite{rebalancingstrategy} is primarily designed for classification, we adapt the process for our specific regression task of trajectory forecasting:
\begin{equation}
\small
d_{i,e} = min(L_F(i,e)-L_F(i,e-1),0)ln(\dfrac{L_F(i,e)}{L_F(i,e-1)}),
\label{d}
\end{equation}
\begin{equation}
\small
a_{i,e} = max(L_{F}(i,e)-L_F(i,e-1),0)ln(\dfrac{L_{F}(i,e)}{L_F(i,e-1)}),
\label{a}
\end{equation}
where $L_{F}(i,e)$ is the loss for instance $i$ at epoch $e$. To label an instance as difficult after $N_{c}$ epochs of observing $a_{i,e}$ and $d_{i,e}$, we  count the number of epochs in which $d_{i,e}>a_{i,e}$, meaning that the training loss has been descending for that instance. The cumulative sum of this indicator over $N_{c}$ epochs is calculated as follows:
\begin{equation}
\small
Count(i)=\sum\limits_{e=1}^{N_c}
\mathbb{I}(d_{i,e}>a_{i,e}).
\label{eq:count}
\end{equation}
An instance is flagged as difficult if $Count(i) < D \times N_{c}$, where $D$ is a predefined threshold parameter.

During the training process, we initially focus on training the social forecaster and social reconstructor across the entire dataset for a fixed number of epochs, $N_T$. This step is important in establishing a robust foundation to enable reasonable reconstructions of difficult cases as part of a continuous training process.
 
The reconstructed version of the samples that are determined to be challenging , its reconstructed masked sequence, $\tilde{S}_o$, are then processed through the social forecaster module to predict future timesteps. We then concatenate the reconstructed sequences with their corresponding forecasted segments as follows:
\begin{equation}
\small
S^{Aug} = \tilde{S_{p}}\oplus SF(\tilde{S_{p}}),
\end{equation}
where $S^{Aug}$ is the newly augmented scene, and $SF(.)$ is the social forecaster function. The integration of identifying and addressing difficult samples occurs in tandem with ongoing training, enhancing the model's capacity to effectively handle such cases without interrupting the training process. This continuous loop ensures that our model dynamically adapts and improves its performance on challenging instances within the training set.

\section{Experiments}
\noindent\textbf{Datasets.} 
We evaluate our method using the ETH/UCY benchmark~\cite{eth,ucy} and SDD\cite{sdd}.
The ETH/UCY benchmark features real world pedestrian trajectories across five scenes including: ETH, Hotel, Univ, Zara1 and Zara2. Each scene contains both multiple and single trajectories, with the multiple trajectory scenarios demonstrating collision avoidance and group behaviour. We use the same coordinate format originally used by SGAN~\cite{socialgan}. For SDD, following DAG-NET~\cite{dagnet} and Social Implicit~\cite{socialimplicit}, we convert the pedestrian locations from pixels to metric coordinates.

\noindent \textbf{Evaluation Metrics.} We use the following three metrics to evaluate our method.

\noindent\textit{Average Distance Error ($\text{ADE}_{\mathsf{K}}$)}: This metric calculates the average $L_{2}$ distance between the ground truth and each of $\mathsf{K}$ trajectories predicted by the model. The formulas for minimum and mean ADE calculation are defined as follows:
\begin{equation}
\small
\text{ADE}^{min}_{\mathsf{K}}= \frac{1}{T}\min_{{k}=1}^{\mathsf{K}}\sum_{t=1}^T\|\hat{S}_f^{k}(n,t) - S_{f}(n,t)\|^2,
\label{ademin}
\end{equation}
\begin{equation}
\small
\text{ADE}^{mean}_{\mathsf{K}}= \frac{1}{\mathsf{K}T}\sum_{k=1}^{\mathsf{K}}\sum_{t=1}^T\|\hat{S}_f^{k}(n,t) - S_{f}(n,t)\|^2.
\label{ademean}
\end{equation}

\noindent \textit{Final Distance Error ($\text{FDE}_{\mathsf{K}}$)}: This metric measures the distance between the predicted final destination and the ground truth final coordinates for all $\mathsf{K}$ predicted trajectories. The formulas for minimum and mean FDE are defined as:
\begin{equation}
\small
    \text{FDE}^{min}_{\mathsf{K}}= \min_{{k}=1}^{\mathsf{K}}\|\hat{S}_f^{k}(n,T) - S_{f}(n,T)\|^2,
    \label{fdemin}
\end{equation}
\begin{equation}
\small
    \text{FDE}^{mean}_{\mathsf{K}}= \frac{1}{\mathsf{K}}\sum_{k=1}^{\mathsf{K}}\|\hat{S}_f^{k}(n,T) - S_{f}(n,T)\|^2.
    \label{fdemean}
\end{equation}

Since all benchmarks use metric measurements, the values of $\text{ADE}_{\mathsf{K}}$ and $\text{FDE}_{\mathsf{K}}$ are expressed in meters. 

\noindent \textit{Kernel Density Estimate-based Negative Log Likelihood (KDE-NLL)}: This metric, which we refer to as KDE, calculates the mean negative log likelihood of the ground truth under the predicted distribution. The distribution is created by fitting a kernel density estimate to trajectory samples~\cite{kde1,kde2}. As KDE measures the negative log likelihood, it is unitless.

\begin{table}[t]
\centering
\setlength{\tabcolsep}{2pt}
\scriptsize
\caption{$\text{ADE}^{min}_{20}\!\!\downarrow/\text{FDE}^{min}_{20}\!\!\downarrow$ (top) and $\text{KDE}\!\!\downarrow$ (bottom) results for the proposed method, and state-of-the-art on the ETH/UCY benchmarks with ${\mathsf{K}}\!=\!20$.}
\begin{tabular}{ccccccc}
\specialrule{1pt}{1pt}{1pt}
\textbf{Method}& 
\textbf{ETH} & 
\textbf{Hotel} &
\textbf{Univ} & 
\textbf{Zara1} & 
\textbf{Zara2} & \textbf{Avg}\\ 

\midrule
\multirow{2}{4em}{SGAN~\cite{socialgan}}  & 
0.63/1.00 & 0.42/0.80 & 0.50/1.01 & 0.29/0.58 & 0.27/0.46  & 0.42/0.77  \\ 
  & 9.388  &  6.328  &  10.557 &  3.484 & 2.639 & 6.479 \\
[0.5ex]
\multirow{2}{*}{S-STGCNN~\cite{socialstgcnn}} & 
0.64/0.93 & 0.33/0.55 & 0.43/0.75 & 0.30/0.50 & 0.26/0.44  & 0.39/0.63  \\ 
  & \underline{2.857}  &  1.215  &  3.395 &  1.590 & 0.955 & 2.002 \\
[0.5ex]
\multirow{2}{*}{S-Implicit~\cite{socialimplicit}}  & 
0.66/1.34 & 0.17/0.31 & 0.29/0.56 & 0.24/0.47 & 0.21/0.42 & 0.31/0.62  \\ 
  & 7.067  &  \underline{0.366}  &  \underline{1.258} &  0.855 &  0.486 & 2.006 \\
 [0.5ex]
\multirow{2}{*}{YNet~\cite{ynet}}  & 
\textbf{0.41}/\textbf{0.53} & \textbf{0.12}/\textbf{0.16} & \underline{0.27}/\underline{0.49} & \underline{0.19}/\underline{0.31} & \underline{0.15}/\underline{0.27}  & \textbf{0.23}/\underline{0.35}\\ 
  & 4.613  &  1.864  &  2.283 &  2.250 & 0.904 & 2.383\\
[0.5ex]
\multirow{2}{*}{Agentformer~\cite{agentformer}}  & 
\textbf{0.45}/\underline{0.75} & \underline{0.14}/0.22 & \textbf{0.25/0.45} & \textbf{0.18/0.30} & \textbf{0.14/0.24}  & \textbf{0.23}/\textbf{0.39}\\ 
  & 3.563  &  0.789  &  1.439 &  \underline{0.793} & \underline{-0.380} & \underline{1.241}\\
[0.5ex]
\multirow{2}{*}{Ours}  & 
0.46/\underline{0.75} & \textbf{0.12}/\underline{0.20} & 0.28/0.51 & 0.21/0.39 & 0.16/0.29& \underline{0.25}/0.43  \\ 
  & \textbf{2.789}  &  \textbf{-0.531}  &  \textbf{1.031} &  \textbf{0.344} & \textbf{-0.704} & \textbf{0.586}\\

\bottomrule
\label{table:stoch}
\end{tabular}
\end{table}

\noindent \textbf{Implementation Details.} 
We follow a leave-one-out strategy used in previous works~\cite{socialgan, agentformer, socialimplicit, socialstgcnn}, where we train our model on 4 sets of scenes, and evaluate on the remaining set. We observe the trajectories for 3.2 seconds (8 timesteps) and predict the future trajectories over the next 4.8 seconds (12 timesteps). 
Following~\cite{agentformer}, we center crop each scene, and apply a random rotation in the range of 0 to 360 degrees.  

Our model contains 8 attention heads in every encoder and decoder module. We train the model on an A100 Nvidia GPU using Adam \cite{adam} with a dropout of 0.1 and a learning rate of 1e-4. All models are trained for 100 epochs. We used grid search to tune our hyperparameters. For more training and architectural details, see Appendix A2.

\section{Results}\label{sect:results}
\noindent \textbf{Performance.} We compare our method against several models, comprising Social Implicit~\cite{socialimplicit}, Social GAN~\cite{socialgan}, Social STGCNN~\cite{socialstgcnn}, YNet~\cite{ynet}, and Agentformer~\cite{agentformer}. The $\text{ADE}^{min}_{20}$/$\text{FDE}^{min}_{20}$ and KDE results for ${\mathsf{K}}\!\!=\!\!20$ model predictions on ETH/UCY are presented in Table~\ref{table:stoch}. Looking at the $\text{ADE}^{min}_{20}$/$\text{FDE}^{min}_{20}$ results, our method shows competitive performance compared to previous methods. However, relying solely on Eqs.~\ref{ademin} and \ref{fdemin}, which are `best of $\mathsf{K}$' metrics, for evaluations may in some cases be misleading. While these metrics evaluate how well one of the produced set of $\mathsf{K}$ samples falls close to the ground truth, it could also improve merely by a method producing a set of trajectories with a large variance that spans across numerous possible future scenarios, and then simply choosing that trajectory that best matches the ground truth. In practice, during inference, ground truth is unknown, and so such a metric could lead us to a low-confidence prediction.

\begin{table}[t]
\setlength{\tabcolsep}{2pt}
\centering
\caption{$\text{ADE}_{1}\!\!\downarrow/\text{FDE}_{1}\!\!\downarrow$ results on the ETH/UCY benchmarks.}
\resizebox{1\columnwidth}{!}{
\begin{tabular}{ccccccc}
\toprule
\textbf{Method}& 
\textbf{ETH} & 
\textbf{Hotel} &
\textbf{Univ} & 
\textbf{Zara1} & 
\textbf{Zara2} & \textbf{Avg}\\ 
\midrule
SGAN~\cite{socialgan} & 1.06/2.14 & 0.61/1.35 & 0.80/1.61 & 0.50/1.07 & 0.46/1.00 & 0.69/1.43  \\

S-STGCNN~\cite{socialstgcnn} & 
1.23/2.12 & 0.61/1.21 & 0.72/1.38 & 0.54/1.08 & 0.46/0.90 &  0.71/1.34 \\ 

S-Implicit~\cite{socialimplicit}   & 
1.06/2.21 & \underline{0.29}/\underline{0.55} & \textbf{0.58/1.22} & \textbf{0.46}/0.99 & 0.41/0.83  &  0.56/1.16  \\
Ynet~\cite{ynet} & 
\textbf{1.00}/\textbf{1.89} & 0.35/0.72 & 0.81/1.79 & 0.52/1.10 & 0.45/1.02  &  0.63/1.30   \\
Agentformer~\cite{agentformer} &
1.06/2.11 & 0.60/1.31 & 0.77/1.62 & 0.79/1.67 & 0.54/1.15&  0.75/1.57 \\
Ours  & 
\textbf{1.00}/\underline{2.00} & \textbf{0.28/0.54} & \underline{0.63}/\underline{1.30} & \textbf{0.46}/\textbf{0.98} & \textbf{0.35/0.76} &  \textbf{0.54}/\textbf{1.12}  \\ 
\bottomrule
\label{table:k1}
\end{tabular}
}
\end{table}

\begin{table}[t]
\centering
\setlength{\tabcolsep}{12pt}
\scriptsize
\caption{$\text{ADE}^{min}_{20}\!\!\downarrow, \text{FDE}^{min}_{20}\!\!\downarrow$ (top) and $\text{KDE}\!\!\downarrow$ (bottom) results for the proposed method, and state-of-the-art on the SDD dataset.}
\begin{tabular}{cccc}
\toprule
\textbf{Method}& 
\textbf{$\text{ADE}_{\text{20}}^{\text{min}}$} & 
\textbf{$\text{FDE}_{\text{20}}^{\text{min}}$} &
\textbf{KDE} \\ 
\midrule
DAG-NET~\cite{dagnet}  & 
0.53 & 1.04 & 1.76 \\
S-Implicit~\cite{socialimplicit} & 
0.47 & 0.89 & 3.89 \\
Ours & 
\textbf{0.33} & \textbf{0.57} & \textbf{0.743} \\
\bottomrule
\label{table:sdd}
\end{tabular}
\end{table}
\begin{table*}[t]
\centering
\setlength
\tabcolsep{7pt}
\scriptsize
\caption{Ablation studies of $\text{ADE}^{min}_{20}\!\downarrow/\text{FDE}^{min}_{20}\!\downarrow$ (top) and KDE$\downarrow$ (bottom) on the ETH/UCY benchmarks.}
\begin{tabular}{cccccccccc}
\specialrule{1pt}{1pt}{1pt}
\multirow{3}{*}{\textbf{Social Recon.}}&\multicolumn{3}{c}{\textbf{Pseudo-trajectory}}&\multirow{3}{*}{\textbf{ETH}} & \multirow{3}{*}{\textbf{Hotel}} &\multirow{3}{*}{\textbf{Univ}} & \multirow{3}{*}{\textbf{Zara1}} & \multirow{3}{*}{\textbf{Zara2}} & \multirow{3}{*}{\textbf{Avg}}\\
\cmidrule{2-4}
~&\textbf{Difficulty based} &\textbf{Random} &\textbf{Inverse} &~ &~ &~ &~& ~  \\


\midrule
\multirow{2}{*}{\cmark} & \multirow{2}{*}{\cmark} & \multirow{2}{*}{\xmark} & \multirow{2}{*}{\xmark}  & 
\textbf{0.46}/\textbf{0.75} & \textbf{0.12}/\textbf{0.20} & \textbf{0.28}/\textbf{0.51} & \textbf{0.21}/\textbf{0.39} & \textbf{0.16}/\textbf{0.29} &
\textbf{0.25}/\textbf{0.43} \\ 
~ & ~ & ~ & ~ & \textbf{2.789} & \textbf{-0.531} & \textbf{1.031} & \textbf{0.344} & \textbf{-0.704} & \textbf{0.586} \\ 
 [0.5ex]
\multirow{2}{*}{\cmark}  & \multirow{2}{*}{\xmark} & \multirow{2}{*}{\cmark} & \multirow{2}{*}{\xmark} & 
0.47/0.79 & 0.13/0.21 & \textbf{0.28}/0.51  & 0.23/0.41 & 0.17/0.30 & 0.26/0.44\\
~ & ~ & ~ & ~ & 2.933 & -0.438 & 1.103& 0.445 & -0.405 & 0.728\\ 
  [0.5ex]
\multirow{2}{*}{\cmark} & \multirow{2}{*}{\xmark} & \multirow{2}{*}{\xmark} & \multirow{2}{*}{\cmark}  &
0.48/0.80 & 0.13/0.23 & \textbf{0.28}/0.53 & 0.22/0.41 & 0.17/0.30 & 0.26/0.45  \\ 
~ & ~  & ~ & ~ & 2.968 & -0.216 & 1.140 & 0.396 & -0.526 & 1.091\\ 
\multirow{2}{*}{\cmark}  & \multirow{2}{*}{\xmark} & \multirow{2}{*}{\xmark} & \multirow{2}{*}{\xmark}  &
0.49/0.80 & 0.13/0.21 & 0.28/0.52 & \textbf{0.21}/0.39 & 0.16/0.29 & \textbf{0.25}/0.44 \\
~ & ~ & ~ & ~ & 2.978 & -0.467 & 1.357 & 0.358 & -0.486 & 0.748\\ 
\multirow{2}{*}{\xmark}  & \multirow{2}{*}{\xmark} & \multirow{2}{*}{\xmark} & \multirow{2}{*}{\xmark} &
0.50/0.83 & 0.13/0.21 & 0.30/0.53 & 0.22/0.41 & 0.17/0.31 & 0.26/0.46 \\
~ & ~ & ~ & ~ & 3.178 & -0.351 & 1.431 & 0.465 & -0.469 & 0.851\\ 
\bottomrule
\label{table:ablation1}
\end{tabular}
\end{table*}

Following previous works such as~\cite{trajectron++}, to address this issue, we use KDE which penalizes the distributions for both inaccuracy and spread.
Unlike the other metrics, KDE considers both the closeness to the true value, as well as the degree of dispersion of a set of hypothesized solutions. In this way, KDE simultaneously rewards accurate and less dispersed distributions,
which together we denote as \emph{stability}. 
The KDE results for our proposed method in Table~\ref{table:stoch} show an improvement (i.e. increased stability) over the state-of-the-art and previous works. Additionally, to further address the issue of `best of $\mathsf{K}$' predictions, we evaluate the methods based on ${\mathsf{K}}\!=\!1$ and rely on $\text{ADE}_{1}$/$\text{FDE}_{1}$. We present these results in Table \ref{table:k1}, where we observe that our method generates stronger predictions in the majority of cases when compared to prior works.
From the results presented in Table~\ref{table:stoch}, we notice that 
Agentformer~\cite{agentformer} uses all of the scenes in the benchmark regardless of the number of pedestrians in the scene. In contrast, Social Implicit \cite{socialimplicit}, Social GAN \cite{socialgan}, Social STGCNN~\cite{socialstgcnn}, and YNet~\cite{ynet} train and test their methods on the same benchmark but exclude scenes with only a single pedestrian. To conduct a fair comparison, we retrained all methods excluding Agentformer~\cite{agentformer} to include both the single pedestrian and multi-pedestrian scenes. This improved over the original reported results in Social STGCNN~\cite{socialstgcnn}, Social GAN~\cite{socialgan} and Social Implicit~\cite{socialimplicit}.
We also evaluate our method on SDD and depict the results in Table~\ref{table:sdd}. Our results outperform the state-of-the-art on this dataset for all three metrics.

\begin{table}[t]
\centering
\setlength
\tabcolsep{3pt}
\caption{Ablation studies of $\text{ADE}^{min}_{20}\!\downarrow/\text{FDE}^{min}_{20}\!\downarrow$ (top) and KDE$\downarrow$ (bottom) on the ETH/UCY benchmark. SL and SA denote Social Loss and Social Attention, respectively.}
\resizebox{0.9\columnwidth}{!}{
\begin{tabular}{cccccccccccc}
\toprule
\textbf{Method}&\multicolumn{1}{c}{\textbf{ETH}} & \multicolumn{1}{c}{\textbf{Hotel}} & \multicolumn{1}{c}{\textbf{Univ}} & \multicolumn{1}{c}{\textbf{Zara1}} & \multicolumn{1}{c}{\textbf{Zara2}} & \multicolumn{1}{c}{\textbf{Avg}}\\
\midrule
\multirow{2}{*}{Ours}  &
\textbf{0.46/0.75} & \textbf{0.12/0.2} & 0.28/\textbf{0.51} & 0.21/0.39 & \textbf{0.16/0.29} & \textbf{0.25/0.43}\\ 
~&2.789 & \textbf{-0.531} & 1.031 & 0.344 & \textbf{-0.704} & \textbf{0.586}
\vspace{3 pt}
\\ 
\multirow{2}{*}{Ours w/o SL}  & 
0.47/0.76 & 0.13/0.21 & 0.28/0.51 & \textbf{0.20/0.38} & 0.17/0.31 & \textbf{0.25/0.43}\\
~&\textbf{2.721} & -0.111 & 1.188 & \textbf{0.310} & -0.264 & 0.769
\vspace{3 pt} \\
Ours w/o SL \& w/o SA 
& 
0.47/\textbf{0.75} & 0.14/0.21 & \textbf{0.26}/0.52 & 0.22/0.42 & \textbf{0.16}/0.31 & \textbf{0.25}/0.44 \\
& 3.086 & -0.51 & \textbf{0.753} & 0.399 & -0.591 & 3.137 \\
\bottomrule
\label{table:ablation4}
\end{tabular}
}
\end{table}

\noindent \textbf{Ablation Studies.} To investigate the contribution of each component in our proposed method, we perform extensive ablation studies on the ETH/UCY benchmark in Table~\ref{table:ablation1}. 
Specifically, we investigate 
the effect of five different scenarios: social reconstruction with three types of augmentation (difficulty-based sampling (ours), random sampling, and inverse sampling),
social reconstruction without augmentation,
and no social reconstruction. For random sampling augmentation, the sequences from the training set are chosen randomly and added to the dataset. For inverse sampling augmentation, we modified Eq.~\ref{eq:count} to $Count(i)=\sum\limits_{e=1}^{N_c}
\mathbb{I}(d_{i,e}<a_{i,e})$. This modification implies selecting the easy sequences in the training data and further augmenting the dataset with these samples. 
We also ablate the augmentation and reconstruction steps in the respective second last and last row of Table~\ref{table:ablation1}.
All the ablated models in the last four rows result in comparable or worse $\text{ADE}^{min}_{20}$/$\text{FDE}^{min}_{20}$ metrics compared to the proposed method. 
The results in Table~\ref{table:ablation1} show our method to outperform all other variations for all datasets. 
\begin{table}[t]

\centering
\setlength
\tabcolsep{2pt}
\caption{Count and percentage of overlap between pedestrians. SL and SA denotes Social Loss and Social Attention, respectively.}
\resizebox{1\columnwidth}{!}{
\begin{tabular}{ccccccccccc}
\toprule
\multirow{2}{*}{\textbf{Method}}&\multicolumn{2}{c}{\textbf{ETH}} & \multicolumn{2}{c}{\textbf{Hotel}} & \multicolumn{2}{c}{\textbf{Univ}} & \multicolumn{2}{c}{\textbf{Zara1}} & \multicolumn{2}{c}{\textbf{Zara2}} \\

~ & Count & \% & Count & \% & Count & \% & Count & \% & Count & \% \\ 
\midrule
Ours & \textbf{38} & \textbf{0.098} & \textbf{297} & \textbf{0.078} & 6667 & 0.007 & \textbf{802} & \textbf{0.074} & \textbf{3548} & \textbf{0.077}\\ 
Ours w/o SL &
48 & 0.124 & 344 & 0.086 & \textbf{6469} & \textbf{0.007} & 850 & 0.075 & 4092 & 0.089 \\
Ours w/o SL \& w/o SA &
66 & 0.170 & 757 & 0.200 & 15172 & 0.018 & 1415 & 0.132 & 12071 & 0.261  \\
\bottomrule
\label{table:count}
\end{tabular}
}
\end{table}




We examine the effect of social loss and social attention in our proposed method in Table~\ref{table:ablation4}. Our proposed method resulted in either the best or second best performance in all three evaluation metrics for all five datasets.
To investigate the effect of social loss on the number of overlaps between forecasted pedestrian locations, we perform ablation studies on the elimination of both social attention and social loss in Table~\ref{table:count}. Eliminating social loss resulted in an increase of overlaps between pedestrians in all subsets of ETH/UCY benchmark except for Univ. We observed that test data for the Univ subset consists of overlaps between locations, possibly due to noise and/or non-birdseye viewpoint recordings of scenes. The viewpoint of non-birdseye videos can result in overlapping pedestrians which do not occupy the same physical space, which does not adhere to the definition of social used here, as discussed in Sec.~\ref{sec:problem}.

\begin{table}[t]
\centering
\caption{$\text{ADE}^{min}_{20}\downarrow/\text{FDE}^{min}_{20}\downarrow$ (top) and KDE$\downarrow$ (bottom) on ETH/UCY for different augmentation and training strategies.}
\resizebox{0.93\columnwidth}{!}{
\setlength\tabcolsep{3pt}

\begin{tabular}{ccccccc}
\toprule
\textbf{Method}&\textbf{ETH} & \textbf{Hotel} &\textbf{Univ} & \textbf{Zara1} & \textbf{Zara2} & \textbf{Avg}\\

\midrule
\multirow{2}{*}{w/ pretrained recon.} &
0.50/0.83 & 0.13/0.22 & 0.29/0.52 & 0.22/0.42 & 0.17/0.32 & 0.26/0.46  \\
~ & 2.845 & -0.140 & 1.203 & 0.474 & -0.412 & 0.794 \\
[0.5ex]
\multirow{2}{*}{w/ initial aug.} &
0.5/0.84 & 0.13/\textbf{0.20} & 0.29/0.52 & \textbf{0.21}/0.39 & 0.17/0.30 & 0.32/0.45 \\
~ & 3.684 & -0.440 & 1.198 & 0.370 & -0.269 & 0.909 \\
[0.5ex]
~&
0.50/0.84 & 0.13/0.21 & 0.30/0.53 & 0.22/0.40 & 0.17/0.30 & 0.26/0.46\\
\multirow{-2}{*}{w/ linear aug. 1} & 3.468 & -0.412 & 1.207 & 0.520 & -0.530 & 0.851\\
[0.5ex]
~&
0.48/0.8 & 0.13/0.21 & 0.29/0.53 & 0.21/0.40 & 0.17/0.30 & \textbf{0.25}/0.45\\
\multirow{-2}{*}{w/ linear aug. 2} & 3.150 & -0.464 & 1.185 & 0.398 & -0.230 & 0.808\\
[0.5ex]
~&
0.50/0.80 & 0.13/0.20 & 0.3/0.55 & 0.21/0.39 & \textbf{0.16}/0.30 & 0.30/0.45\\
\multirow{-2}{*}{w/ social force aug.} & 2.948 & -0.478 & 1.299 & 0.421 & -0.317 & 0.775\\
[0.5ex]
Ours &
\textbf{0.46}/\textbf{0.75} & \textbf{0.12}/\textbf{0.20} & \textbf{0.28}/\textbf{0.51} & \textbf{0.21}/\textbf{0.38} & \textbf{0.16}/\textbf{0.29} & \textbf{0.25}/\textbf{0.43}\\
~ & \textbf{2.789} & \textbf{-0.531} & \textbf{1.031} & \textbf{0.344} & \textbf{-0.704} & \textbf{0.586}\\
\bottomrule
\label{table:ablation2}
\end{tabular}
}
\end{table}

As depicted earlier in Figure~\ref{fig:method} the generated pseudo-trajectories evolve alongside the rest of the framework and are fed back to the model during subsequent training cycles. Here, we aim to investigate the impact of this strategy based on several variants of our model. First instead of allowing the social reconstructor to continue to train alongside the rest of the model, we pre-train and freeze it in our framework. We refer to this variant as `w/ pretrained recon'. Next, in the second variant, we train and then place the social reconstructor outside of our model as a serial data augmentation module. This variant is referred to as `w/ initial aug' in the table. We then explore three additional variants where we augment the data and train the social forecaster using the augmented data. We extrapolate the trajectories linearly based on the first two or the last two timesteps in the past horizon. We call these two augmentation methods `linear aug 1' and `linear aug 2', respectively. As the final augmentation, we generate samples inspired by the social force model~\cite{simulation}. We add an equal number of samples for all augmentations as our original model (ETH: 1,616, Hotel: 1,493, Univ: 1,277, Zara1: 1,365, Zara2: 370). This investigation shows that the  $\text{ADE}^{min}_{20}$/$\text{FDE}^{min}_{20}$ and KDE values deteriorate as a result of the mentioned modifications further demonstrating the benefits of our strategy in co-training the social reconstructor and forecaster, and re-using the augmented trajectories during training.

\begin{figure}[t]
\centerline{\includegraphics[width=0.96\columnwidth]{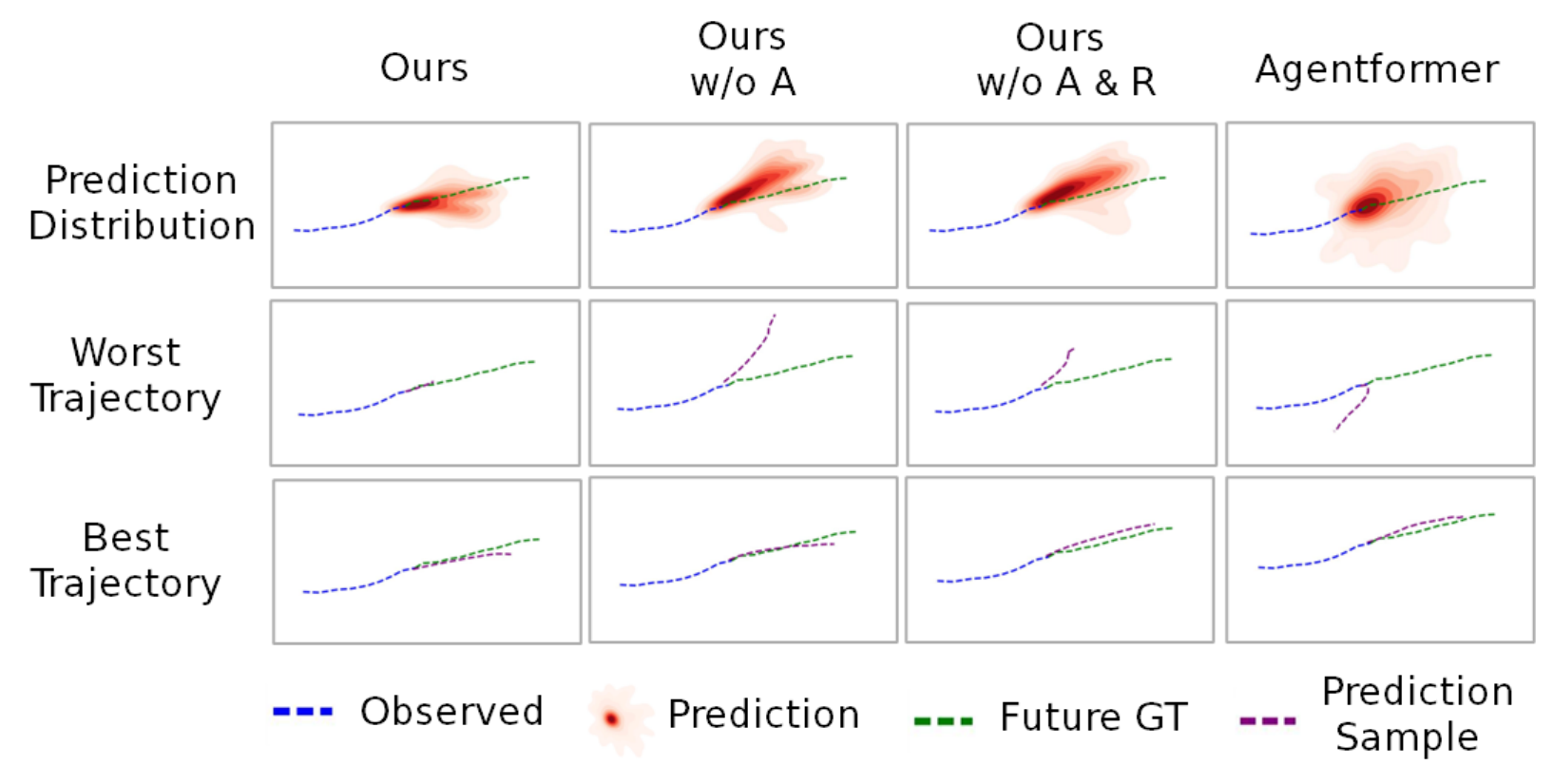}}
\caption{Prediction results for our method compared to two ablated models as well as Agentformer on two examples of the ETH scene. 
Past and future ground truth trajectories are shown in blue and green dashed lines, while the prediction samples are illustrated with purple dashed lines. `Ours w/o A' indicates our method without Augmentations. `Ours w/o A and R' is our method without the augmentations and social reconstructor. }


\label{fig:tra}
\end{figure}


\noindent\textbf{Discussion.} 
We provide a qualitative comparison of our method with two ablated versions as well as Agentformer on two example scenes from the ETH test set in Figure~\ref{fig:tra}. The second column shows our proposed method without augmented data, while the third presents an ablation without both augmentation and the social reconstructor. Looking at the prediction distribution, our method performs more accurately and with less variance among the predicted trajectories compared to other ablated varieties. The accuracy of the distribution also shows itself in the comparison among worst trajectories. The worst trajectories in our method are the best among all the ablations, while the best trajectories are comparable. This also indicates a limitation of the $\text{ADE}^{min}_{20}$/$\text{FDE}^{min}_{20}$ evaluation metric, which only favors the best results. Our method also outperforms Agentformer qualitatively when comparing the prediction distributions and the worst trajectories, for which Agentformer produces more disperse and less directionally focused trajectories. More examples and comparison with an ablation of our method without the social loss are included in Appendix A.3.

\begin{figure}[t]
\centerline{\includegraphics[width=1\columnwidth]{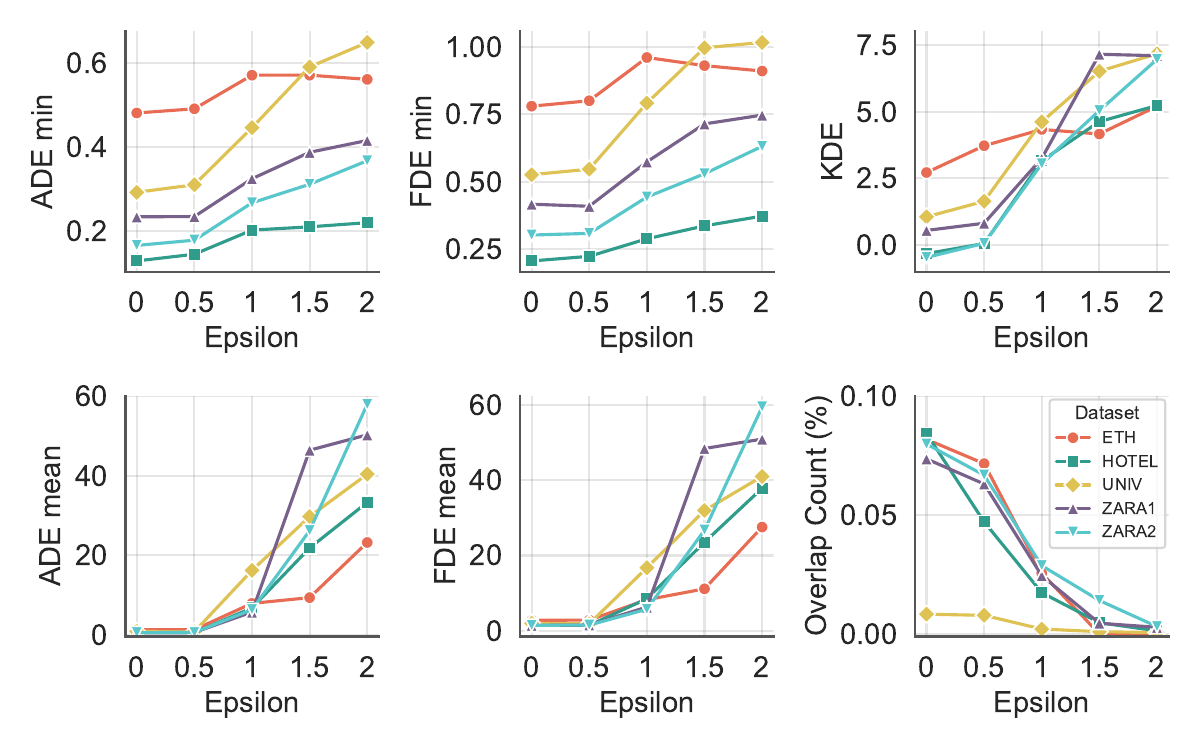}}
\caption{Sensitivity analysis on the value of $\epsilon$ and its effect on six evaluation metrics: $\text{ADE}^{min}_{20}$, $\text{FDE}^{min}_{20}$, $\text{ADE}^{mean}_{20}$, $\text{FDE}^{mean}_{20}$, KDE, and the overlap count percentage between pedestrians.}
\label{fig:epsilon_sensitivity}
\end{figure}

\begin{figure}[t!]
\centerline{\includegraphics[width=0.95\columnwidth]{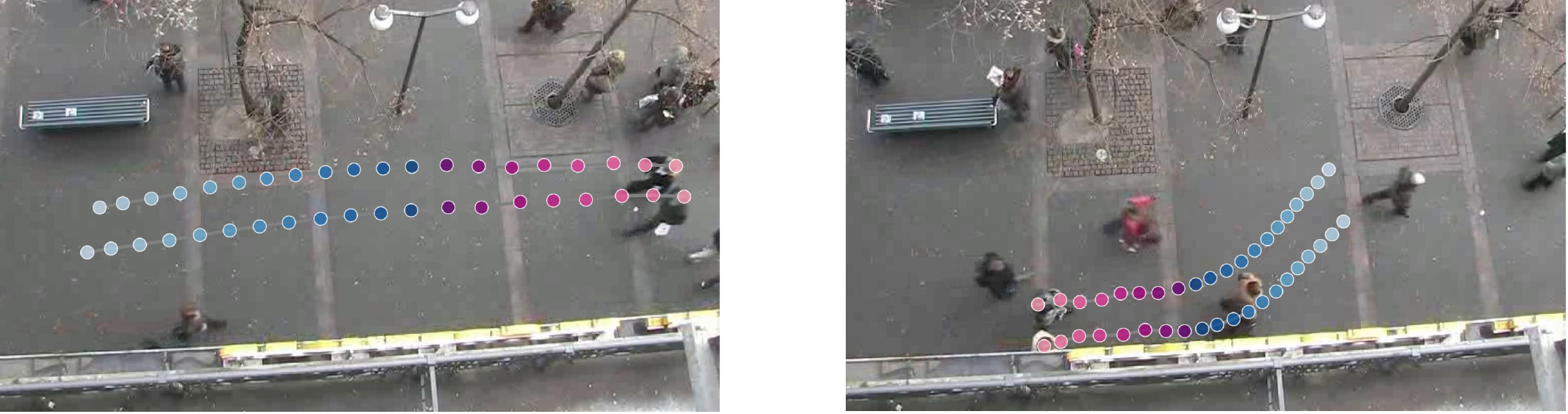}}
\caption{Examples of trajectory prediction with targets in close proximity of each other. Past and future timesteps are denoted by red and blue colors, respectively.
}
\label{fig:proximity_main}
\end{figure}

To investigate the effect of the minimal accepted distance between pedestrians $\epsilon$ (Sec.~\ref{sec:problem}), as well as the number of overlaps between predicted pedestrian locations, we executed a sensitivity analysis on ETH/UCY dataset, where we increased $\epsilon$  from 0 to 2 meters by increments of 0.5. As illustrated in Figure~\ref{fig:epsilon_sensitivity}, an increase from 0 to 0.5 does not result in a significant change in 
$\text{ADE}^{min}_{20}\!/\text{FDE}^{min}_{20}\!$  and $\text{ADE}^{mean}_{20}\!/\text{FDE}^{mean}_{20}\!$. However, further increasing it results in a significant deterioration of the mentioned evaluation metrics. KDE also degrades and the overlap between forecasted pedestrians further decreases with an increase in $\epsilon$. Our social loss ensures that pedestrians maintain a minimum distance of $\epsilon$. As per Eqs. \ref{eq:socf} and \ref{eq:socr}, using higher values for this hyperparameter will result in enforced distances between pedestrians in the scene, which can artificially increase the minimum separation between predictions, thus negatively effecting performance.
Finally, to further study the impact of our social loss on the distance between pedestrians walking side-by-side, we explore a number of examples of such cases to identify any possible adverse effects. Figure~\ref{fig:proximity_main} presents two examples, where we observe consistent predictions for the two target trajectories. Additional samples are provided in Appendix A.3.

\section{Conclusion}

In this paper, we introduced a
pedestrian trajectory forecaster that uses a social loss to generate socially-informed pseudo-trajectories.  
Our proposed method uses a reconstructor to generate pseudo-trajectories which are used to augment the learning process. We have shown that injecting these pseudo-trajectories from the partially trained network improves results when compared to augmenting from a similar statically generated offline dataset. Our novel loss function decreases the number of overlaps  between predicted pedestrian locations at future timesteps, while keeping the predictions accurate and increasing stability of predictions. Our experiments on standard benchmark datasets and metrics show our method to outperform existing state-of-the-art trajectory prediction methods.

\small
\bibliographystyle{ieee_fullname}
\bibliography{egbib}

\begin{thebibliography}{10}\itemsep=-1pt

\bibitem{sociallstm}
Alexandre Alahi, Kratarth Goel, Vignesh Ramanathan, Alexandre Robicquet, Li Fei-Fei, and Silvio Savarese.
\newblock Social lstm: Human trajectory prediction in crowded spaces.
\newblock In {\em IEEE/CVF Conference on Computer Vision and Pattern Recognition}, pages 961--971, 2016.

\bibitem{textaug}
Markus Bayer, Marc-Andr{\'e} Kaufhold, and Christian Reuter.
\newblock A survey on data augmentation for text classification.
\newblock {\em ACM Computing Surveys}, 55(7):1--39, 2022.

\bibitem{onboard}
Apratim Bhattacharyya, Mario Fritz, and Bernt Schiele.
\newblock Long-term on-board prediction of people in traffic scenes under uncertainty.
\newblock In {\em IEEE/CVF Conference on Computer Vision and Pattern Recognition}, pages 4194--4202, 2018.

\bibitem{contextaware}
Haleh Damirchi, Michael Greenspan, and Ali Etemad.
\newblock Context-aware pedestrian trajectory prediction with multimodal transformer.
\newblock In {\em IEEE International Conference on Image Processing}, pages 2535--2539, 2023.

\bibitem{imaug1}
Alhussein Fawzi, Horst Samulowitz, Deepak Turaga, and Pascal Frossard.
\newblock Adaptive data augmentation for image classification.
\newblock In {\em International Conference on Image Processing}, pages 3688--3692, 2016.

\bibitem{socialgan}
Agrim Gupta, Justin Johnson, Li Fei-Fei, Silvio Savarese, and Alexandre Alahi.
\newblock Social gan: Socially acceptable trajectories with generative adversarial networks.
\newblock In {\em IEEE/CVF Conference on Computer Vision and Pattern Recognition}, pages 2255--2264, 2018.

\bibitem{simulation}
Dirk Helbing and Peter Molnar.
\newblock Social force model for pedestrian dynamics.
\newblock {\em Physical review E}, 51(5):4282, 1995.

\bibitem{kde2}
Boris Ivanovic and Marco Pavone.
\newblock The trajectron: Probabilistic multi-agent trajectory modeling with dynamic spatiotemporal graphs.
\newblock In {\em IEEE/CVF International Conference on Computer Vision}, pages 2375--2384, 2019.

\bibitem{adam}
Diederik~P Kingma and Jimmy Ba.
\newblock Adam: A method for stochastic optimization.
\newblock {\em arXiv preprint arXiv:1412.6980}, 2014.

\bibitem{ucy}
Alon Lerner, Yiorgos Chrysanthou, and Dani Lischinski.
\newblock Crowds by example.
\newblock In {\em Computer Graphics Forum}, volume~26, pages 655--664, 2007.

\bibitem{ynet}
Karttikeya Mangalam, Yang An, Harshayu Girase, and Jitendra Malik.
\newblock From goals, waypoints \& paths to long term human trajectory forecasting.
\newblock In {\em IEEE/CVF International Conference on Computer Vision}, pages 15233--15242, 2021.

\bibitem{notthejourney}
Karttikeya Mangalam, Harshayu Girase, Shreyas Agarwal, Kuan-Hui Lee, Ehsan Adeli, Jitendra Malik, and Adrien Gaidon.
\newblock It is not the journey but the destination: Endpoint conditioned trajectory prediction.
\newblock In {\em European Conference on Computer Vision}, pages 759--776, 2020.

\bibitem{socialstgcnn}
Abduallah Mohamed, Kun Qian, Mohamed Elhoseiny, and Christian Claudel.
\newblock Social-stgcnn: A social spatio-temporal graph convolutional neural network for human trajectory prediction.
\newblock In {\em IEEE/CVF Conference on Computer Vision and Pattern Recognition}, pages 14424--14432, 2020.

\bibitem{socialimplicit}
Abduallah Mohamed, Deyao Zhu, Warren Vu, Mohamed Elhoseiny, and Christian Claudel.
\newblock Social-implicit: Rethinking trajectory prediction evaluation and the effectiveness of implicit maximum likelihood estimation.
\newblock In {\em European Conference on Computer Vision}, pages 463--479, 2022.

\bibitem{dagnet}
Alessio Monti, Alessia Bertugli, Simone Calderara, and Rita Cucchiara.
\newblock Dag-net: Double attentive graph neural network for trajectory forecasting.
\newblock In {\em International Conference on Pattern Recognition}, pages 2551--2558, 2021.

\bibitem{walkalone}
Stefano Pellegrini, Andreas Ess, Konrad Schindler, and Luc Van~Gool.
\newblock You'll never walk alone: Modeling social behavior for multi-target tracking.
\newblock In {\em IEEE/CVF International Conference on Computer Vision}, pages 261--268, 2009.

\bibitem{eth}
Stefano Pellegrini, Andreas Ess, and Luc Van~Gool.
\newblock Improving data association by joint modeling of pedestrian trajectories and groupings.
\newblock In {\em European Conference on Computer Vision}, pages 452--465, 2010.

\bibitem{pedformer}
Amir Rasouli and Iuliia Kotseruba.
\newblock Pedformer: Pedestrian behavior prediction via cross-modal attention modulation and gated multitask learning.
\newblock In {\em IEEE International Conference on Robotics and Automation}, pages 9844--9851, 2023.

\bibitem{pie}
Amir Rasouli, Iuliia Kotseruba, Toni Kunic, and John~K Tsotsos.
\newblock Pie: A large-scale dataset and models for pedestrian intention estimation and trajectory prediction.
\newblock In {\em IEEE/CVF International Conference on Computer Vision}, pages 6262--6271, 2019.

\bibitem{sdd}
Alexandre Robicquet, Amir Sadeghian, Alexandre Alahi, and Silvio Savarese.
\newblock Learning social etiquette: Human trajectory understanding in crowded scenes.
\newblock In {\em European Conference on Computer Vision}, pages 549--565, 2016.

\bibitem{etiquette}
Alexandre Robicquet, Amir Sadeghian, Alexandre Alahi, and Silvio Savarese.
\newblock Learning social etiquette: Human trajectory understanding in crowded scenes.
\newblock In {\em European Conference on Computer Vision}, pages 549--565, 2016.

\bibitem{survey}
Andrey Rudenko, Luigi Palmieri, Michael Herman, Kris~M Kitani, Dariu~M Gavrila, and Kai~O Arras.
\newblock Human motion trajectory prediction: A survey.
\newblock {\em The International Journal of Robotics Research}, 39(8):895--935, 2020.

\bibitem{trajectron++}
Tim Salzmann, Boris Ivanovic, Punarjay Chakravarty, and Marco Pavone.
\newblock Trajectron++: Dynamically-feasible trajectory forecasting with heterogeneous data.
\newblock In {\em European Conference on Computer Vision}, pages 683--700, 2020.

\bibitem{shomeecg}
Debaditya Shome, Pritam Sarkar, and Ali Etemad.
\newblock Region-disentangled diffusion model for high-fidelity ppg-to-ecg translation.
\newblock In {\em AAAI Conference on Artificial Intelligence}, volume~38, pages 15009--15019, 2024.

\bibitem{behavior}
Jianhua Sun, Qinhong Jiang, and Cewu Lu.
\newblock Recursive social behavior graph for trajectory prediction.
\newblock In {\em IEEE/CVF Conference on Computer Vision and Pattern Recognition}, pages 660--669, 2020.

\bibitem{anomaly}
Qiyue Sun and Yang Yang.
\newblock Unsupervised video anomaly detection based on multi-timescale trajectory prediction.
\newblock {\em Computer Vision and Image Understanding}, 227:103615, 2023.

\bibitem{kde1}
Luca~Anthony Thiede and Pratik~Prabhanjan Brahma.
\newblock Analyzing the variety loss in the context of probabilistic trajectory prediction.
\newblock In {\em IEEE/CVF International Conference on Computer Vision}, pages 9954--9963, 2019.

\bibitem{sgnet}
Chuhua Wang, Yuchen Wang, Mingze Xu, and David~J Crandall.
\newblock Stepwise goal-driven networks for trajectory prediction.
\newblock {\em IEEE Robotics and Automation Letters}, 7(2):2716--2723, 2022.

\bibitem{Autoformer}
Haixu Wu, Jiehui Xu, Jianmin Wang, and Mingsheng Long.
\newblock Autoformer: Decomposition transformers with auto-correlation for long-term series forecasting.
\newblock {\em Advances in Neural Information Processing Systems}, 34:22419--22430, 2021.

\bibitem{imaug2}
Mingle Xu, Sook Yoon, Alvaro Fuentes, and Dong~Sun Park.
\newblock A comprehensive survey of image augmentation techniques for deep learning.
\newblock {\em Pattern Recognition}, 137:109347, 2023.

\bibitem{autonomous}
Bo Yang, Song Yan, Zheng Wang, and Kimihiko Nakano.
\newblock Prediction based trajectory planning for safe interactions between autonomous vehicles and moving pedestrians in shared spaces.
\newblock {\em IEEE Transactions on Intelligent Transportation Systems}, 2023.

\bibitem{bitrap}
Yu Yao, Ella Atkins, Matthew Johnson-Roberson, Ram Vasudevan, and Xiaoxiao Du.
\newblock Bitrap: Bi-directional pedestrian trajectory prediction with multi-modal goal estimation.
\newblock {\em IEEE Robotics and Automation Letters}, 6(2):1463--1470, 2021.

\bibitem{rebalancingstrategy}
Sihao Yu, Jiafeng Guo, Ruqing Zhang, Yixing Fan, Zizhen Wang, and Xueqi Cheng.
\newblock A re-balancing strategy for class-imbalanced classification based on instance difficulty.
\newblock In {\em IEEE/CVF Conference on Computer Vision and Pattern Recognition}, pages 70--79, 2022.

\bibitem{agentformer}
Ye Yuan, Xinshuo Weng, Yanglan Ou, and Kris~M Kitani.
\newblock Agentformer: Agent-aware transformers for socio-temporal multi-agent forecasting.
\newblock In {\em IEEE/CVF International Conference on Computer Vision}, pages 9813--9823, 2021.

\bibitem{tnt}
Hang Zhao, Jiyang Gao, Tian Lan, Chen Sun, Ben Sapp, Balakrishnan Varadarajan, Yue Shen, Yi Shen, Yuning Chai, Cordelia Schmid, et~al.
\newblock Tnt: Target-driven trajectory prediction.
\newblock In {\em Conference on Robot Learning}, pages 895--904, 2021.

\end{thebibliography}

\clearpage

\appendix
\renewcommand\theequation{A.\arabic{equation}}
\renewcommand\thefigure{A.\arabic{figure}}
\renewcommand\thetable{A.\arabic{table}}
\renewcommand\thesection{A.\arabic{section}}

\setcounter{equation}{0}
\setcounter{figure}{0}
\setcounter{table}{0}
\setcounter{section}{0}

\section*{Appendix}
\section{Pseudocode}

Our proposed method is detailed in Algorithm~\ref{alg:SocRec}. We train our model for $N_T$ epochs initially. During this warm-up period, we also record the values of the loss $L_F$ for each sample $i$ and epoch $e$. After this period, we calculate $Count$ for each sample $i$ and determine their inclusion as a pseudo-trajectory if $Count(i)<D\!\times\!N_{c}$. Here, $N_{c}$ denotes the number of epochs where the loss for each sample has been recorded. At the end of warm-up period $N_{c}\!=\!N_{T}$, while after the warm-up period $N_{c}\!=\!N_{Int}$,
where $N_{Int}$ is the epoch interval between pseudo-trajectory generations. To generate the final augmented samples we concatenate the reconstructed past timestep $ \tilde{S}_{p}$ and the social forecaster output trajectory $SF(\tilde{S}_{p})$. 
Prior to each augmentation, we erase the previously added trajectories from the training data.

\section{Training and Architectural Details}

\noindent\textbf{Hyperparameters}
The hyperparameters that we used to train our models with are depicted in Table~\ref{table:hyper}. 
We observed that the Univ dataset was sensitive to overfitting, due to a higher number of test samples compared to the train samples, as well as the difference between the fewer number of crowded scenes in the train partition compared to the larger number in the test partition. 
To effectively address this, the size (number of parameters) of the model was reduced for this dataset, by reducing the values of the hyperparameters $d_m$ and $d_{ff}$, as shown in the first two rows of Table~\ref{table:hyper}. 
We used the Steplr scheduler, which has the two hyperparameters of gamma and stepsize as depicted in Table~\ref{table:hyper}. 

\begin{table}[h]
\centering
\setlength
\tabcolsep{1pt}
\footnotesize
\caption{Hyperparameters of our method for ETH/UCY and SDD.}
\begin{tabular}{cccccccc}
\specialrule{1pt}{1pt}{1pt}
\textbf{Hyper-}&\multicolumn{6}{c}{\textbf{Dataset}}& \multirow{2}{*}{\textbf{Description}}\\
\cmidrule{2-7}
\textbf{Params} & \multirow{1}{*}{\textbf{ETH}} & \multirow{1}{*}{\textbf{Hotel}} & \multirow{1}{*}{\textbf{Univ}} & \multirow{1}{*}{\textbf{Zara1}} & \multirow{1}{*}{\textbf{Zara2}} & \multirow{1}{*}{\textbf{SDD}}\\
\midrule
$d_{m}$ &
128 & 64 & 64 & 256 & 128 & 128 & Model dimension\\ 
\vspace{2pt}
$d_{ff}$ &
512 & 256 & 128 & 512 & 512 & 256 & Feedforw. layer dim. \\
\vspace{2pt}
$d_z$ &
32 & 32 & 32 & 32 & 32 & 32 & Latent space dim.\\
\vspace{2pt}
$n^{f}_{enc}$ &
1 & 2 & 2 & 1 & 2 & 1 & Encoder layers\\
\vspace{2pt}
$n^{f}_{dec}$ &
1 & 1 & 1 & 1 & 1 & 1 & Decoder layers\\
\vspace{2pt}
$n^{r}_{dec}$ &
1 & 1 & 1 & 1 & 1 & 1 & Recon. decoder layers\\
\vspace{2pt}
$n_{atthead}$ &
8 & 8 & 8 & 8 & 8 & 8 & Attention heads\\
\vspace{2pt}
$D$ &
0.5 & 0.5 & 0.5 & 0.5 & 0.5 & 0.5 & Difficulty threshold\\
\vspace{2pt}
$\epsilon$ &
0.1 & 0.1 & 0.05 & 0.1 & 0.1 & 0.1 & Epsilon for social loss\\
\vspace{2pt}
$N_{T}$ &
10 & 20 & 20 & 20 & 20 & 10 & Threshold epoch\\
\vspace{2pt}
$N_{Int}$ &
10 & 10 & 10 & 10 & 10 & 10 & interval epochs\\
\vspace{2pt}
$R$ &
30 & 10 & 10 & 30 & 20 & 10 & Masking ratio \\
\vspace{2pt}
$gamma$ &
0.8 & 0.8 & 0.8 & 0.5 & 0.8 & 0.8 & Steplr scheduler Gamma\\
\vspace{2pt}
$lr$ &
1e-4 & 1e-4 & 1e-4 & 1e-4 & 1e-4 & 1e-4 & Learning rate \\
\vspace{2pt}
$stepsize$ &
10 & 20 & 20 & 10 & 40 & 10 & Step size for scheduler \\
\vspace{2pt}
$w_1$ &
1 & 1 & 1 & 1 & 1  & 1 & Forecaster loss weight \\
\vspace{2pt}
$w_2$ &
1 & 1 & 1 & 1 & 1  & 1 & Recon. loss weight \\
\vspace{2pt}
$w_3$ &
1 & 1 & 1 & 1 & 1  & 1 & Social loss weight \\
\bottomrule
\label{table:hyper}
\end{tabular}
\end{table}

\begin{algorithm}[h]
\small
\caption{Training of our proposed method}\label{alg:SocRec}
$N_{Tot}$: Total number of epochs, 
$N_{T}$: Threshold epoch,\\
$N_{c}$: Loss observation duration,
$N_{Int}$: Interval epoch,\\
$N_{m}$: Number of Training samples,
$N_{a}$: Number of Augmented samples,\\
$SF$: Forecaster module, 
$SR$: Reconstructor module,\\
$S_{p}$: Past trajectory,
$S^{masked}_{p}$: Masked past trajectory,\\
$S_{f}$: Future ground truth trajectory,\\
$L_F$: Forecaster CVAE loss function, \\
$L_R$: Reconstructor VAE loss function,\\
$\mathcal{L}_{Total}$: Total loss, 
$L_{Soc}$: Social loss function,\\
$l_{arr}\in\mathbb{R}^{N_{m} \times N_{c}}$ :Array to save losses,\\
$D$: Difficulty Threshold,\\
$A_{arr}\in\mathbb{R}^{N_{a}}$: Array to save Augmented samples,\\
\While{$e < N_{Tot}$}{
    \While{$i < N_{m}$}{
        $\tilde{S}_{f} \gets SF(S_{p})$;\\
        $\tilde{S}_{p} \gets SR(S^{masked}_{p})$;\\
        Calculate $L_F$, $L_R$, $L_{Soc}$, $\mathcal{L}_{Total}$
        Compute gradients and backpropagate $\mathcal{L}_{Total}$\\
        $l_{arr}[i,e]\gets L_F$\\
        \If{$e = N_{Thr}$}{
            $Count(i) \gets \sum\limits_{e=1}^{N_{thr}}\mathbb{I}(d_{i,e}>a_{i,e})$\\
            \If{$Count(i) < D \times N_{c}$}{
            \BlankLine
            $A_{arr} \gets \tilde{S}_{p} \oplus SF(\tilde{S}_{p})$\\
            } 
        $N_{thr} \gets N_{thr} + N_{Int}$\\
        }
        $i \gets i+1$\\
    }
\If{$e = N_{Thr}$}{
    Erase previously added augmented smaples\\
    Add $A_{arr}$ samples to the training set\\
    Clear $A_{arr}$,$l_{arr}$\\
    }
$e \gets e+1$\\
}
\end{algorithm}

\noindent\textbf{Masking.}
To mask each scene, we calculate the number of total timesteps $T_{scene}=N\!\times\!t_{p}$. The number of masked timesteps can be calculated as $R\!\times\!T_{scene}$ for masking ratio $R$. We observed that masking a timestep solely by setting it to location zero was confusing to the model, as it would get interpreted as a non-masked zero location. For this reason, we concatenated a binary indicator with the masked input location $S^{masked}_{p}(t)$ for each timestep $t$, where $0$ indicates no masking, and $1$ indicates masking.


\section{Additional Visualizations and Results}

\begin{figure}[t!]
\centerline{\includegraphics[width=0.9\columnwidth]{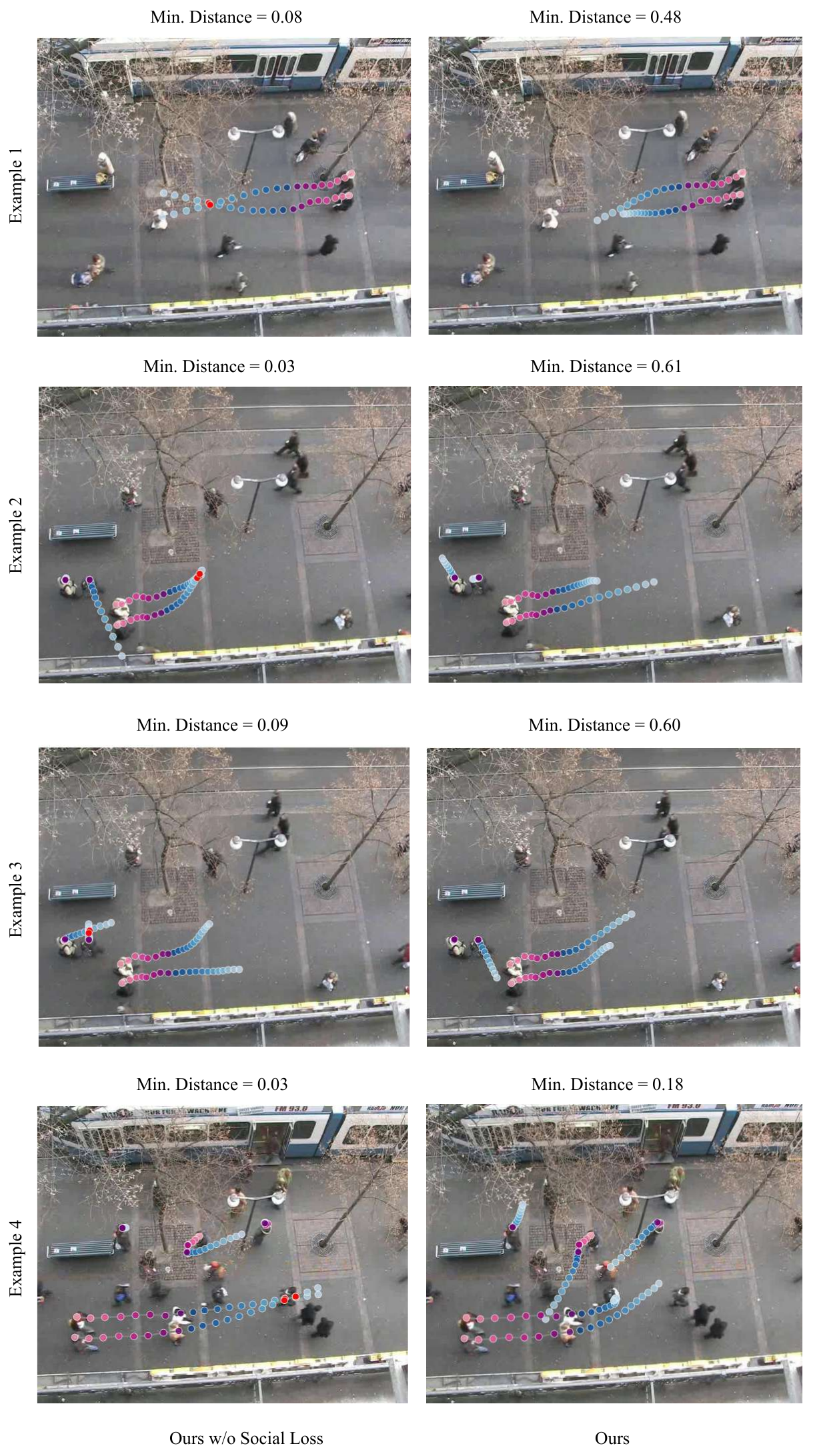}}
\caption{Visualization of trajectories in a scene from Hotel subset. Red circles show location overlaps. Min. Distance denotes the minimum euclidean distance between pedestrians in meters.
Past and future timesteps are denoted by red and blue, respectively.}
\label{fig:vis_1}
\end{figure}

\begin{figure}[t!]
\centerline{\includegraphics[width=0.8\columnwidth]{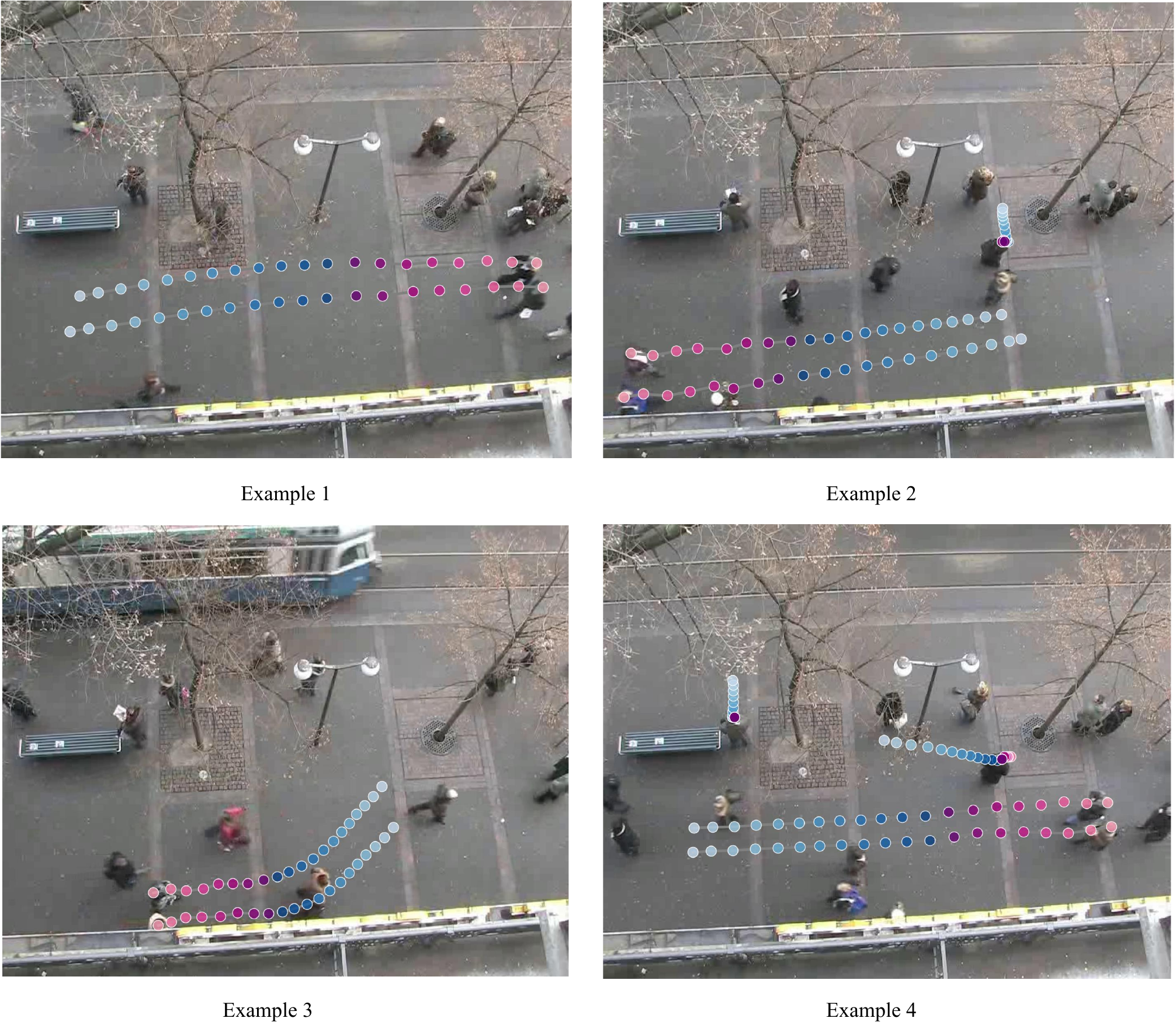}}
\caption{Examples of trajectory prediction with targets in close proximity of each other. Past and future timesteps are denoted by red and blue, respectively.
}
\label{fig:proximity}
\end{figure}

In this section, we provide visualizations of forecasted trajectories for four examples shown in Figure~\ref{fig:vis_1} to illustrate the effect of social loss on the number of location overlaps. 
According to the standard protocol,
trajectories are included within each scene only for those
pedestrians whose trajectories (combining both ground truth past and future locations) comprise a length of 20 timesteps.
For example, in Example 1, there are two such pedestrians, in Example 2, there are four, etc.
The minimum distance between pairs of pedestrian trajectories in the scene is depicted above each example. Overlaps between pedestrians, where their separation within a timestep is smaller than $\epsilon\le0.1$, are highlighted with red circles. 
As shown in the figure, our proposed method provides improved socially-aware predictions, where the forecasted trajectories have a lower chance of overlapping with each other. 
We also investigate the effect of social loss on $\text{ADE}^{mean}_{20}$ and $\text{FDE}^{mean}_{20}$, which is shown in Table~\ref{table:d}. Our method results in better performance regarding the mean error of the produced trajectories in four out of five subsets. More examples illustrating our method's predictions in close proximity cases are shown in Figure~\ref{fig:proximity}. 

\begin{table}[t!]
\centering
\setlength
\tabcolsep{2pt}
\footnotesize
\caption{Ablation studies for $\text{ADE}^{mean}_{20}\downarrow/\text{FDE}^{mean}_{20}\downarrow$.}
\begin{tabular}{ccccccc}
\toprule
\textbf{Social} & \textbf{Social} & \multicolumn{5}{c}{\textbf{Dataset}} \\ 
\textbf{Attention}&\textbf{Loss}&\multicolumn{1}{c}{\textbf{ETH}} & \multicolumn{1}{c}{\textbf{Hotel}} & \multicolumn{1}{c}{\textbf{Univ}} & \multicolumn{1}{c}{\textbf{Zara1}} & \multicolumn{1}{c}{\textbf{Zara2}} 
\\
\midrule
\cmark  & \cmark &
\textbf{1.27/2.61} & \textbf{0.57/1.33} & 0.86/1.89 & \textbf{0.70/1.52} & \textbf{0.59/1.34} \\ 

\cmark  & \xmark &
1.37/2.81 & 0.64/1.40 & 0.89/1.90 & 0.75/1.63 & 0.68/1.51 \\
\xmark  & \xmark &
1.38/2.78 & \textbf{0.57}/1.21 & \textbf{0.82/1.80} & 0.79/1.73 & 0.68/1.52 \\
\bottomrule
\label{table:d}
\end{tabular}
\end{table}

\begin{figure}[!t]
\centerline{\includegraphics[width=0.9\columnwidth]{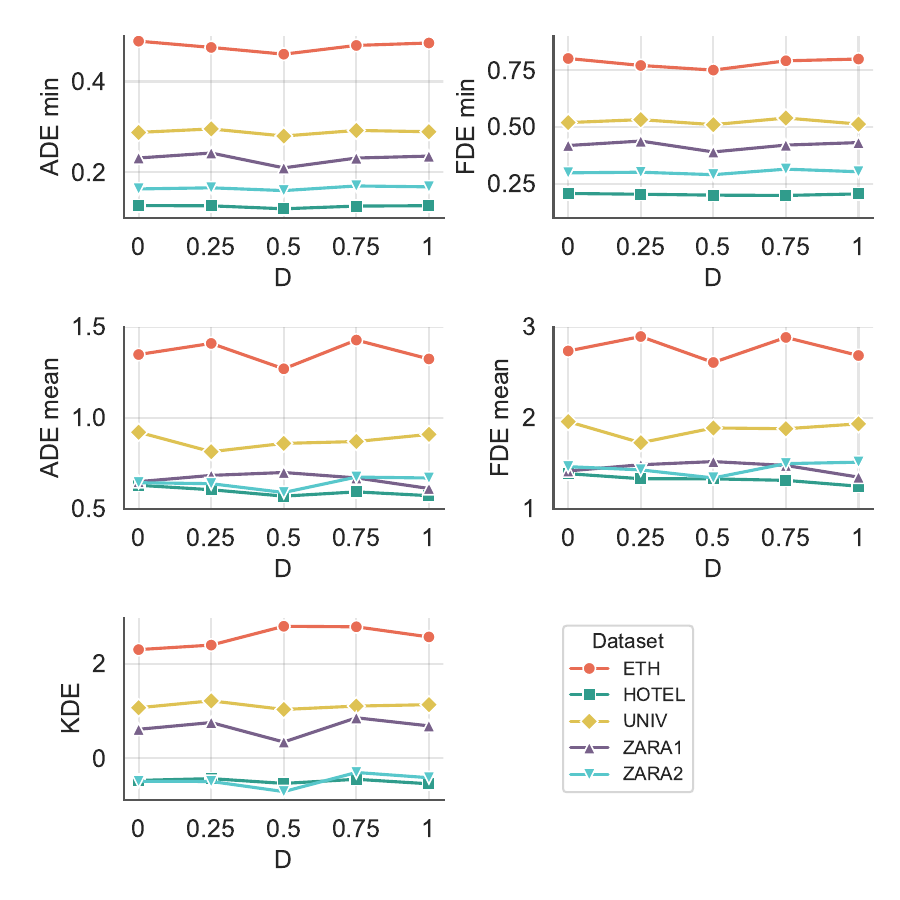}}
\caption{Sensitivity analysis for $D$.}
\label{fig:d_sensitivity}
\end{figure}

To analyze the effect of threshold $D$ on the evaluation metrics, we perform a sensitivity analysis on this hyperparameter for different values between 0 and 1, where increasing $D$ results in the inclusion of augmentations of easier samples in the training data. The results are depicted in Figure~\ref{fig:d_sensitivity}, where we observe that $D=0.5$ achieves the best overall results by effectively balancing the inclusion and exclusion of augmented samples during training.

\begin{figure*}[t]
\centerline{\includegraphics[width=0.90\textwidth]{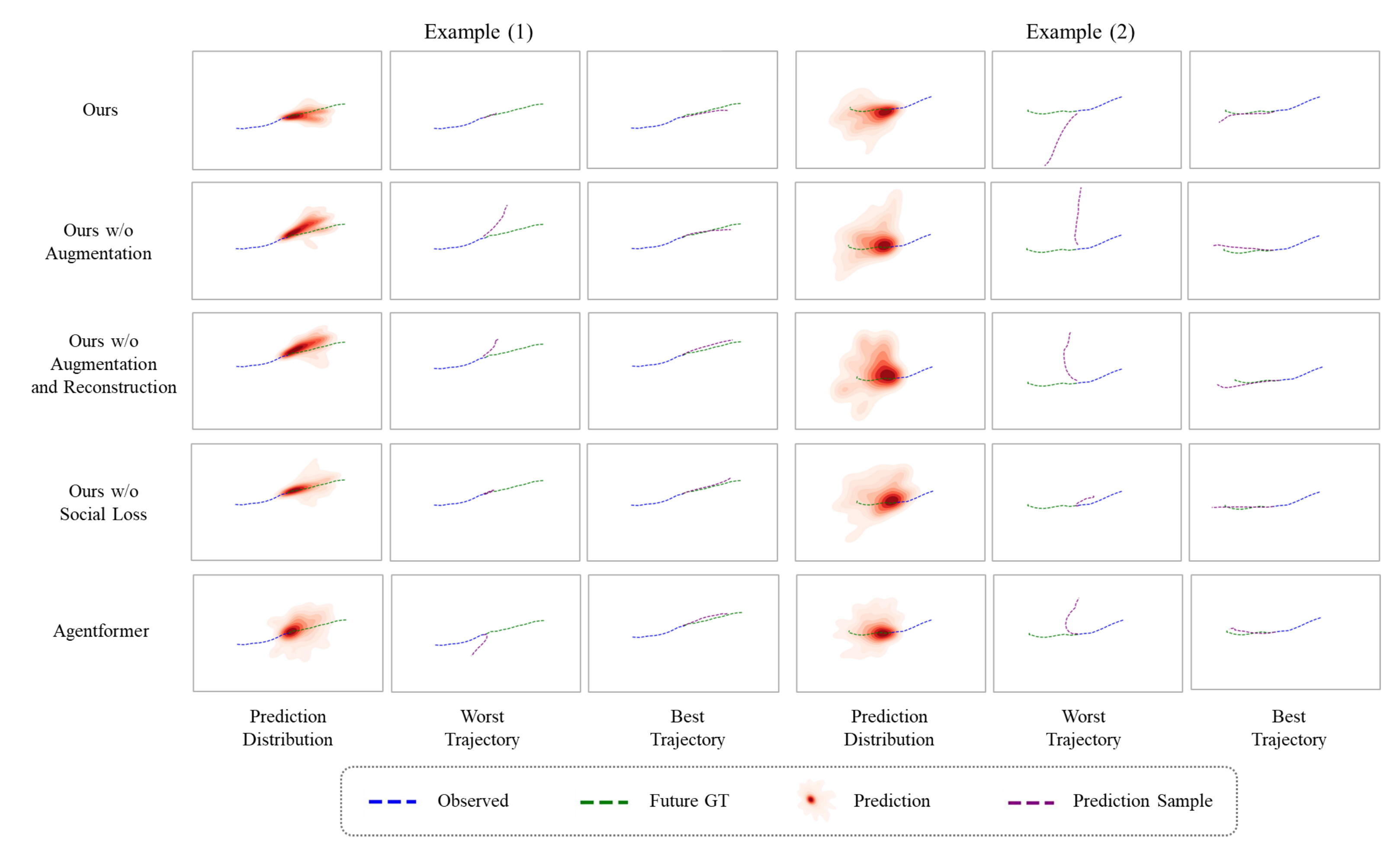}}
\caption{Prediction results for our method compared to three ablated models as well as Agentformer on two examples of the ETH scene. 
Past and future ground truth trajectories are shown in blue and green dashed lines, while the prediction samples are illustrated with purple dashed lines. We observe that our proposed method produces a less dispersed distribution compared to all the ablated versions as well as Agentformer. Our proposed method, compared to the others, also produces the closest `worst trajectories' to the ground truth, while predicting comparable `best trajectories' to others.
}
\label{fig:tra}
\end{figure*}

Two examples demonstrating the best, worst, and distribution of predicted trajectories by our proposed method, three ablated versions of our method, and Agentformer \cite{agentformer} are illustrated in Figure~\ref{fig:tra_supp}. We observe that our method produces less dispersed distributions, as well as better `best case' and more viable `worst case' predictions.

\end{document}